\documentclass{article}

\usepackage[square,sort,comma,numbers]{natbib}
\usepackage{arxiv}
\usepackage{soul}
\usepackage[utf8]{inputenc} 
\usepackage[T1]{fontenc}    
\usepackage{hyperref}       
\usepackage{url}            
\usepackage{booktabs}       
\usepackage{amsfonts}       
\usepackage{nicefrac}       
\usepackage{microtype}      
\usepackage{lipsum}
\usepackage{graphicx}
\title{Convolutional Collaborative Filter Network for Video Based Recommendation Systems}

\author{
  Hsieh, Cheng-Kang
  \and
  Campo, Miguel\thanks{Corresponding author: miguelangel.campo-rembado@fox.com. All authors are with 20th Century Fox. First author is also with Miso Technologies}
  \and
  Taliyan, Abhinav
  \and
  Nickens, Matt
  \and
  Pandya, Mitkumar
  \and
  Espinoza, JJ
}

\begin{document}
\maketitle

\begin{abstract}
This analysis explores the temporal sequencing of objects in a movie trailer. Temporal sequencing of objects in a movie trailer (e.g., a long shot of an object vs intermittent short shots) can convey information about the type of movie, plot of the movie, role of the main characters, and the filmmakers cinematographic choices. When combined with historical customer data, sequencing analysis can be used to improve predictions of customer behavior. E.g., a customer buys tickets to a new movie and maybe the customer has seen movies in the past that contained similar sequences. To explore object sequencing in movie trailers, we propose a video convolutional network to capture actions and scenes that are predictive of customers' preferences. The model learns the specific nature of sequences for different types of objects (e.g., cars vs faces), and the role of sequences in predicting customer future behavior. We show how such a temporal-aware model outperforms simple feature pooling methods proposed in our previous works and, importantly, demonstrate the additional model explain-ability allowed by such a model.

\end{abstract}


\section{Introduction}

Understanding detailed audience composition is important for movie studios that invest in stories of uncertain commercial outcome. One source of uncertainty is that studios do now know how the movie is going to be like--what is it going to feel like--until the last few months before release. The other related source of uncertainty is audience and market fluidity. Studios don't know with certainty 'how', and especially 'which' audiences are going to respond because the movie isn't finished and because the strength and nature of competitive movies is also unknown.

An important function at movie studios is understanding the micro-segmentation of the customer base. E.g., not all super hero movies bring the same audience, etc. Over the last years, studios have invested in data tools to learn and to map out the customer segments, and to make predictions for future films. Granular predictions at the micro-segment level, and even at the customer level, have became routine inputs into important business decisions, and provide a trusted barometer of the potential financial performance of the movie.

Recommendation systems for movie theatrical releases have emerged as valuable tools that are especially well suited to provide granular forward looking audience projections to support greenlighting decisions, movie positioning studies, and marketing and distribution. MERLIN, the recommendation system for theatrical releases built at 20th Century Fox, is used to predict user attendance and segment indexes a year in advance, and to refine the prediction with anonymized user behavior signals.

Predicting user behavior far in advance of the movie release is an example of \textit{pure cold-start} prediction and is challenging for movies that are novel, movies that are non-sequels, and movies that cross traditional genres. Recent research has explored using movie synopses \cite{campo2018collaborative} and movie trailers (\cite{campo2018competitive}, \cite{Lee:2018:CDM:3219819.3219856}), combined with collaborative filter models, to predict which customers consume which movies. In our analysis, Campo et al. showed that recommendations made based on video data are qualitatively different from those based on the synopsis data \cite{campo2018competitive}. This finding can be explained by the different information content of the two media.

An open question when training video-based models is the choice of the feature space. A popular approach due to its simplicity is to analyze, individually, the different frames of a video using a trained image classification deep architecture. With this approach, the dense feature representations of the different frames can be pooled together in a single dense representation through an averaging or max element-wise operation. For example, Campo et al. \cite{campo2018competitive} use the average pooling of image features of individual video frames and use it as the video features.

Video analysis using pooling schemes to collapse an entire video or part of a video into a unique dense feature vector can miss important semantic aspects of the video. Although simple to implement, the approach neglects the sequential and temporal aspects of the story, which, we argue, can be useful for characterization of a motion picture. For example, a trailer with a long close-up shot of a character is more likely for a drama movie than for an action movie, whereas a trailer with quick but frequent shots is more likely for an action movie. In both cases, though, average pooling would result in the same feature vector.

A second question when training video-based models is the level of semantic expressiveness of the model. One could use individual frames as unit of analysis, and apply image classification models to create a dense feature vector for each frame. Those vectors represent the 'objects' depicted in the frames (car, face, etc), and can then be pooled together to create a video-level feature vector. A collaborative filter can use the presence of the objects in the trailer to predict customer behavior (e.g., maybe because the customer has seen movies in the past that also depicted some of those objects in the trailer).

Alternatively, one could try to use the sequences of ordered frames to identify patterns in the presence of objects that repeat themselves multiple times, perhaps at some regular time intervals, and perhaps across different trailers. Some of those sequences could indicate actual actions. For example, intermittent close-up shots of actors faces could be indicative of dialog, whereas intermittent shots of cars could be indicative of a car chase. One could feed those identified sequences, represented in a suitable dense vectorization, to a collaborative filter to determine whether the presence of some sequence in a trailer is predictive of customer behavior (e.g., maybe because the customer has seen movies in the past that also contained such sequence).

In this paper, we explore a temporal-aware model that takes temporal dynamics of movie trailer imagery into account to more faithfully capture salient elements that are conveyed not in individual frames, but over sequences of frames, and that hopefully create a better representation of the elements of the actual story. Our model is based on the idea of \textit{convolution over time} \cite{lecun1995convolutional}. As in previous work, for each video frame, we extract dense image features using a pre-trained image feature extractor. Then, we apply multi-layered convolution filters over the dense image features of multiple consecutive video frames. Convolutional filters are capable of capturing signals from not specific to individual frames, but that result from the combination of a sequence of frames that are in the filter receptive field. A longer receptive field can capture actions that unfold over longer periods of time. The convolution layer is followed by a temporal pooling layer to summarize the signals throughout the video before fed into a hybrid-collaborative filtering (CF) pipeline. The convolutional filters and collaborative filters are trained end-to-end and against millions of moviegoers' attendance records. This allows them to learn video actions (or non-actions) that are most predictive of users' movie preferences.

\begin{figure*}[t]
\centering
  \includegraphics[width=1\linewidth]{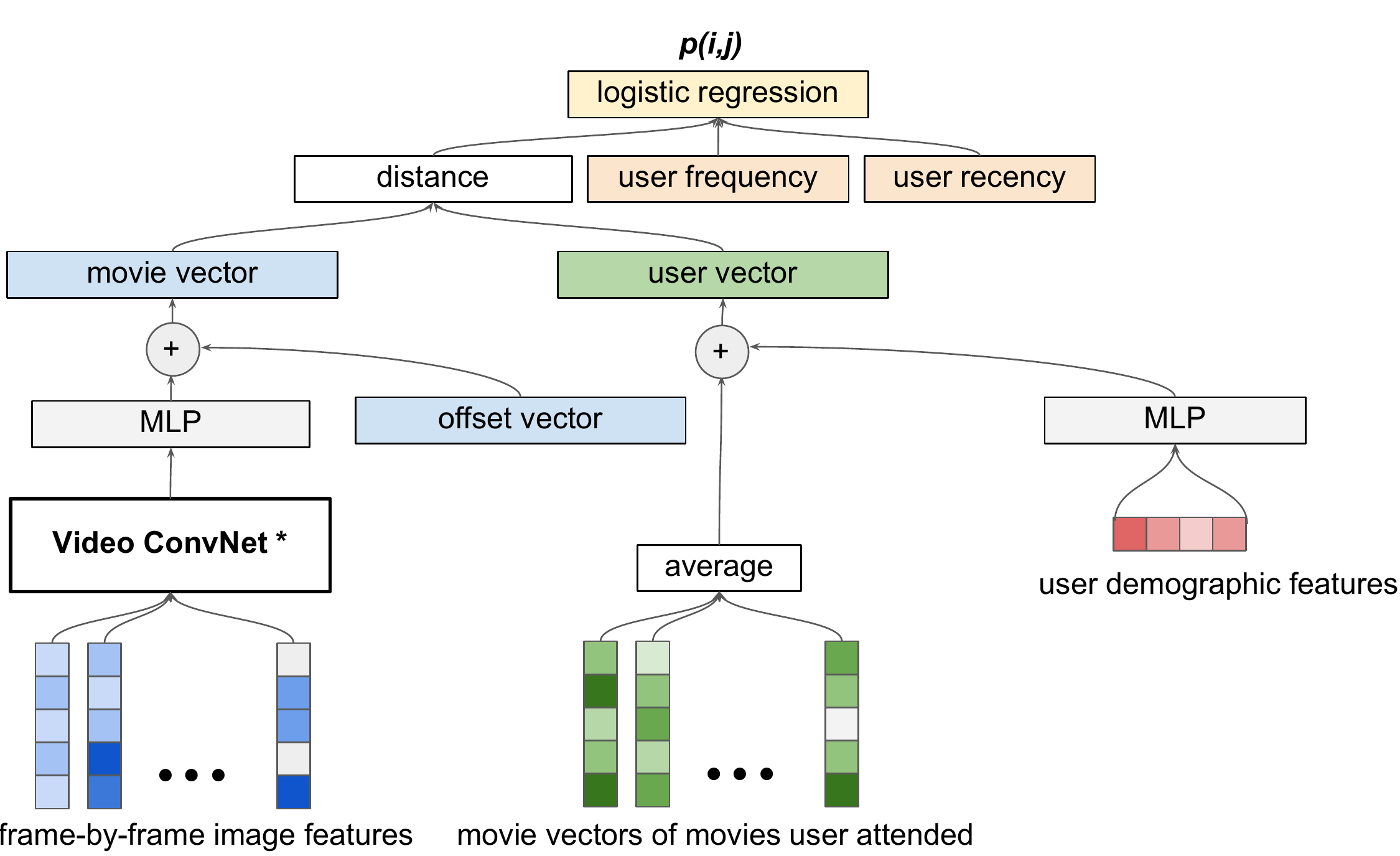}
  \caption{Overview of Merlin's hybrid collaborative filtering pipeline. A logistic regression layer combines a distance-based CF model with user's frequency and recency features to estimate the movie attendance probability. The model is trained end-to-end, and the loss of the logistic regression at the top is backpropogated to every trainable component.}
  \label{fig:overview}
\end{figure*}

\section{Background}
Movie recommendation for online streaming platforms has been well-studied in RecSys literature \cite{youtube, koren2009matrix, harper2016movielens}. However, little research has been done to study the recommendation and prediction problems for theatrical releases. Specifically, the study of the \textit{cold-start} prediction problem before and during movie production \cite{schein2002methods}. This paper is part of a series of works that report the development of \textbf{Merlin}, an experimental movie attendance prediction and recommendation system. 

At its core, Merlin is a hybrid collaborative filtering pipeline that is enabled a fully anonymized, user privacy compliant, movie attendance dataset that combines data from different sources with hundreds of movies released over the last years, and millions of attendance records. Figure \ref{fig:overview} shows a high-level overview of Merlin. Each movie in Merlin is modeled by a fixed-dimensional vector (referred to as \textit{movie vector} in the figure) that are extracted from either movie synopsis or movie trailer data. Each user in Merlin is modeled by a user vector that is the sum of the movie vectors of the movies she attended, and the features from her basic demographics information.

The focus of this paper is at the bottom left of the figure, a video convolutional network architecture that is in charge of learning the components of the movie vector from the analysis of sequences in the video trailers. Please refer to \cite{campo2018competitive} for a detailed description of the remaining part of the pipeline.

\begin{figure*}[t]
\centering
  \includegraphics[width=0.5\linewidth]{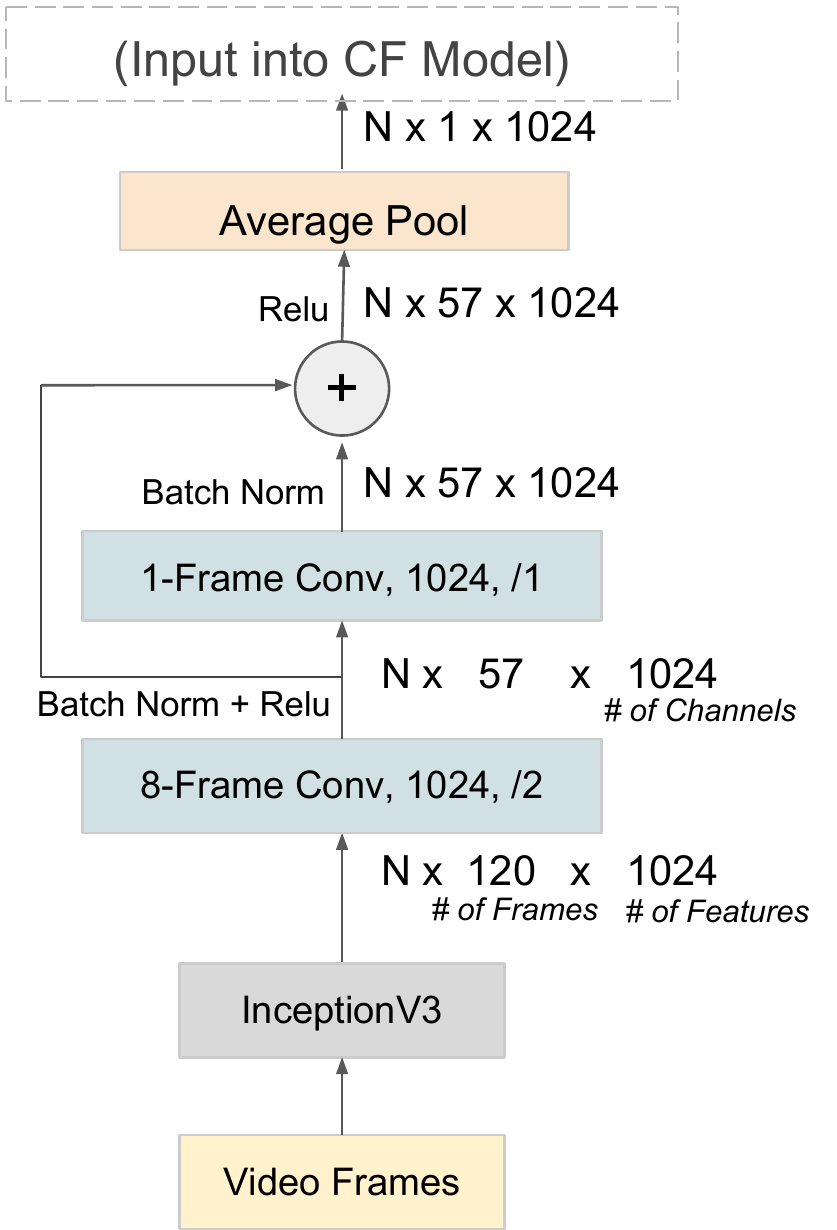}
  \caption{Video Convolutional Model for Trailers. $N$ denotes the number of trailers. Each trailer consists of 120 image frames (1 fps). 1024-dim image features are extracted by a pre-trained Inception V3 \cite{szegedy2016rethinking}. A convolutional layer with 1024, 8-frame-conv, filters is applied, where each filter is of the shape (8 x 1024) and is applied over the temporal (i.e. frame) dimension. Next, a residual conv layer with 1024, 1-frame-conv, filters is applied to increase the model capacity, followed by an average pooling layer to summarize the signals throughout a trailer. }
  \label{fig:arch}
\end{figure*}

\section{Video Convolutional Model for Movie Trailers}
The high level idea of the proposed video convolutional model is to learn a collection of filters, each of which captures a particular kind of object-sequence that could be suggestive of specific \textit{actions}. For example, a pair of filters may learn a sequence of images of a country road and a car, which could be suggestive of someone driving down the country road; another pair of filters could learn intermittent sequences of a car and a person, which could indicate someone driving aggressively on the street and being chased, etc. 

Assuming that certain object-sequences are universal across different movie trailers, and that different actions (and storylines) follow distinct \textit{object-sequence templates} (for example, a complex car chasing action may ensue a car chase, followed by a car flipping, and a car explosion), the job of the network is to fit a \textit{object-specific temporal convolutional filters} to learn such a object-sequence templates.

There are countless varieties of object-sequences that would appear in a movie trailer, and we will not have sufficient amount of data to learn all of them. Our approach is to learn those that are relevant in the prediction problem on the the customer transactional data.  This is why we let movie attendance data guide the convolutional filters to focus on those actions that are most predictive of users' preferences.

Figure \ref{fig:arch} illustrates the proposed network structure. Specifically, the raw input to our video convolutional model are video frames extracted from the movie trailers. We first down-sample the videos to 1 frame-per-second, and only use the first 120 seconds of the videos as the input data (i.e. each trailer has 120 frames of data). We use a pre-trained Inception V3 model \cite{szegedy2016rethinking} to extract 1024 dimensional image features for each frame. A convolution layer with 1024 convolutional filters are applied against image features. Each filter is of the shape (8 x 1024), where 8 is the filter size (i.e. 8-frame) and 1024 is the size of the input channel (i.e. the number of image features). This layer alone has about 8 million parameters (i.e. 1024 x 8 x 1024) and is capable of mixing information from 8 video frames. We apply these filters with stride = 2 along the temporal-dimension without padding, which reduces the size of the temporal dimension from 120 to 57.

Then, we apply a residual layer that performs another set of convolutional filters against the output of the previous layer. If this layer's filter size > 1-frame, it will further increase the effective receptive field. However, in our experiment (see Section \ref{sec:result}), we found that an 1-frame convolution performs the best. This 1-frame convolution layer consists of 1024 filters, each of which has the shape (1 x 1024). Such filters do not expand the receptive field, but still increase the model capacity by mixing information across all the 1024 input channels \cite{lin2013network}, which is equivalent to applying a \textit{fully-connected layer} to each input frame. In our case, this layer seems to learn to exploit the \textit{inter-correlation} between object-sequences. For example, a gun fighting sequences may have a high correlation with a explosion scene. This 1-frame convolution layer may exploit such correlation and provide additional evidences (or counter evidences) to the occurrence of certain object-sequences. 

Finally, an average pooling is applied to summarize the signals across the temporal dimension of a trailer. As shown in Figure \ref{fig:overview}, the output of average pooling will go through a multi-layer perceptron (MLP) before being used as the movie vector for the corresponding movie. As described earlier, the whole pipeline (including the video convolution part and CF part) is trained end-to-end to minimize the prediction error for user attendance. By doing this, we force the convolution filters to focus on those actions that are most predictive of users' preferences.

\section{Performance Evaluation}
We evaluate the proposed model using a fully anonymized, user privacy compliant, movie attendance dataset with millions of attendance records from hundreds of movies released over the last years. Among the latest 300 movie release, we hold out the attendance records of the most recent 50 movies for cold-start evaluation to simulate the prediction accuracy prior to the movie release. For the remaining 250 movies, we random sample 80\% of their attendance records for training, and hold out 10\% for validation and another 10\% for testing. All the models are trained until convergence on the validation set and evaluated on the testing set. 

The model is trained using stochastic gradient descent with mini-batches. Every batch contains an mix of even number of positive and negative user-movie pairs. A positive user-movie pair represents that a user went to that particular movie according to our records, while a negative pair consists of a user and a movie randomly sampled from all the movies that user did not go to. For evaluation, we sample five million user-movie pairs with a 1-to-9 positive-negative ratio to simulate the common average movie attendance rate.

\subsection{Evaluation Results}\label{sec:result}
We compared the proposed video convolutional model to the following baselines: 1) \textbf{Merlin + Text} that uses movie synopses vectorized via a word2vec model as movie features \cite{campo2018competitive}. 2) \textbf{Merlin + Video AvgPool} a temporal-unaware model that applies AvgPool to collapse the image features of every video frame in a trailer. To have a fair comparison, we apply the same video-preprocessing steps, including downsampling to 1-fps and limiting the trailer length to 120 frames, to both \textbf{Merlin + Video AvgPool} and the proposed \textbf{Merlin + Video Convolution}.

We use Area-Under-Curve (AUC) \cite{hanley1982meaning} as our performance metric. As our ultimate goal is to determine what kind of moviegoers a movie will attract, and provide insights for movie production, AUC serves as an indicator to the overall prediction power of the models, which is more suitable than the per-user ranking metrics, such as Top-k recall.

\begin{table}[t]
\centering
\label{table:perf}
\begin{tabular}{|l|c|c|}
\cline{1-3}
\textbf{Model} & \textbf{In Matrix} & \textbf{Cold Start} \\ 
Merlin + Text            &   0.849            &  0.731   \\
Merlin + Video AvgPool            &    0.845               &       0.723              \\
Merlin + Video Convolution            &      0.849         &    0.751        \\ \cline{1-3}
\end{tabular}
\caption{Area-Under-Curve of Different Models}
\end{table}

Table \ref{table:perf} summarizes the evaluation results. As shown, every model shows comparable performance for the \textit{in-matrix} movies, where the existing attendance records are available. This is not surprising because the in-matrix accuracy is mostly determined by the collaborative filtering part of the model, which is the same across all three models. However, the models differ in their \textit{cold-start} performances (i.e. for new movies that do not have attendance records yet). Specifically, the proposed \textbf{Merlin + Video Convolution} outperforms the \textbf{Merlin + Text} and  \textbf{Merlin + Video AvgPool} by 2 and almost 3 absolute percentage points respectively. The fact that the proposed convolution model outperforms the AvgPool model suggests that the proposed convolutional architecture is a more effective way to extract video features that are predictive of users' preferences. Moreover, the fact that our video-based model is able to surpass a text-based model hints a new research avenue into utilizing more and richer multimedia contents to improve the movie recommendations.
\begin{figure*}[t]
\centering
  \includegraphics[width=0.7\linewidth]{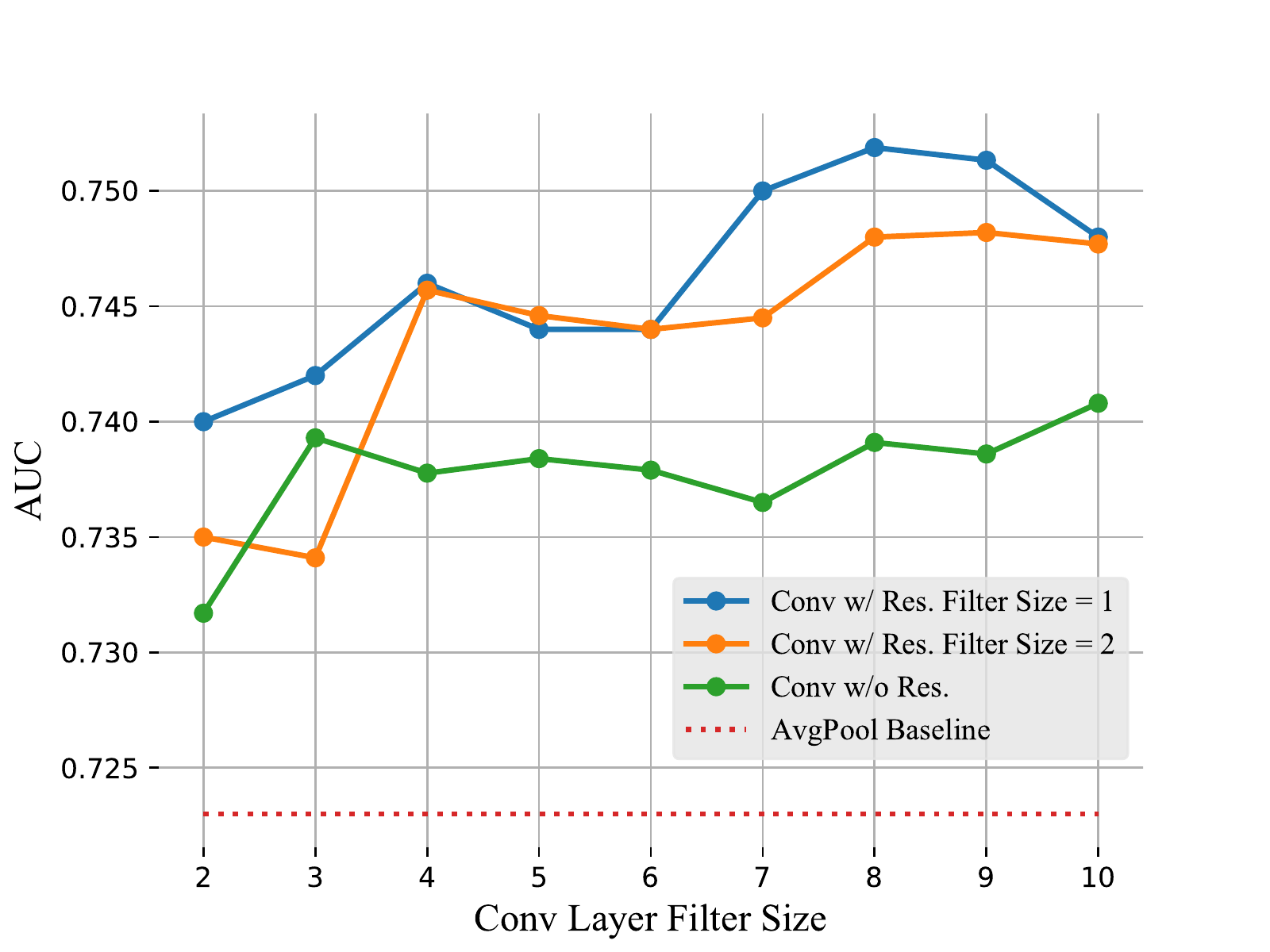}
  \caption{Cold-start performance of varying filter sizes and different model architecture. The results suggest that (1) a sufficiently-large receptive field (e.g. 8 frames) is crucial to the prediction accuracy, and (2) the proposed residual layer is beneficial, in particular, when 1-frame convolution is used.}
  \label{fig:results}
\end{figure*}

\subsection{Length of Sequence Study}
One main argument of this paper is the importance of object-sequences in video understanding. We hypothesize that a convolutional network needs a sufficiently-large receptive field to capture different object-sequences in videos. Figure \ref{fig:results} shows the performance of varying filter sizes and different convolutional network settings to shed some light on this hypothesis. We make two observations: (1) a sufficiently-large filter size (e.g. larger than 4 frames) is crucial to the model performance, and the performance starts to saturate when the filter sizes are larger than 8 frames; and (2) the residual layer is beneficial to the model performance, in particular, when 1-frame convolution is used in the residual layer. This might suggest that the positive effect of residual layer is due more to its capability of exploiting inter-correlation between different scenes than to its capability of expanding receptive fields.

\begin{figure*}[t]
\centering
  \includegraphics[width=1\linewidth]{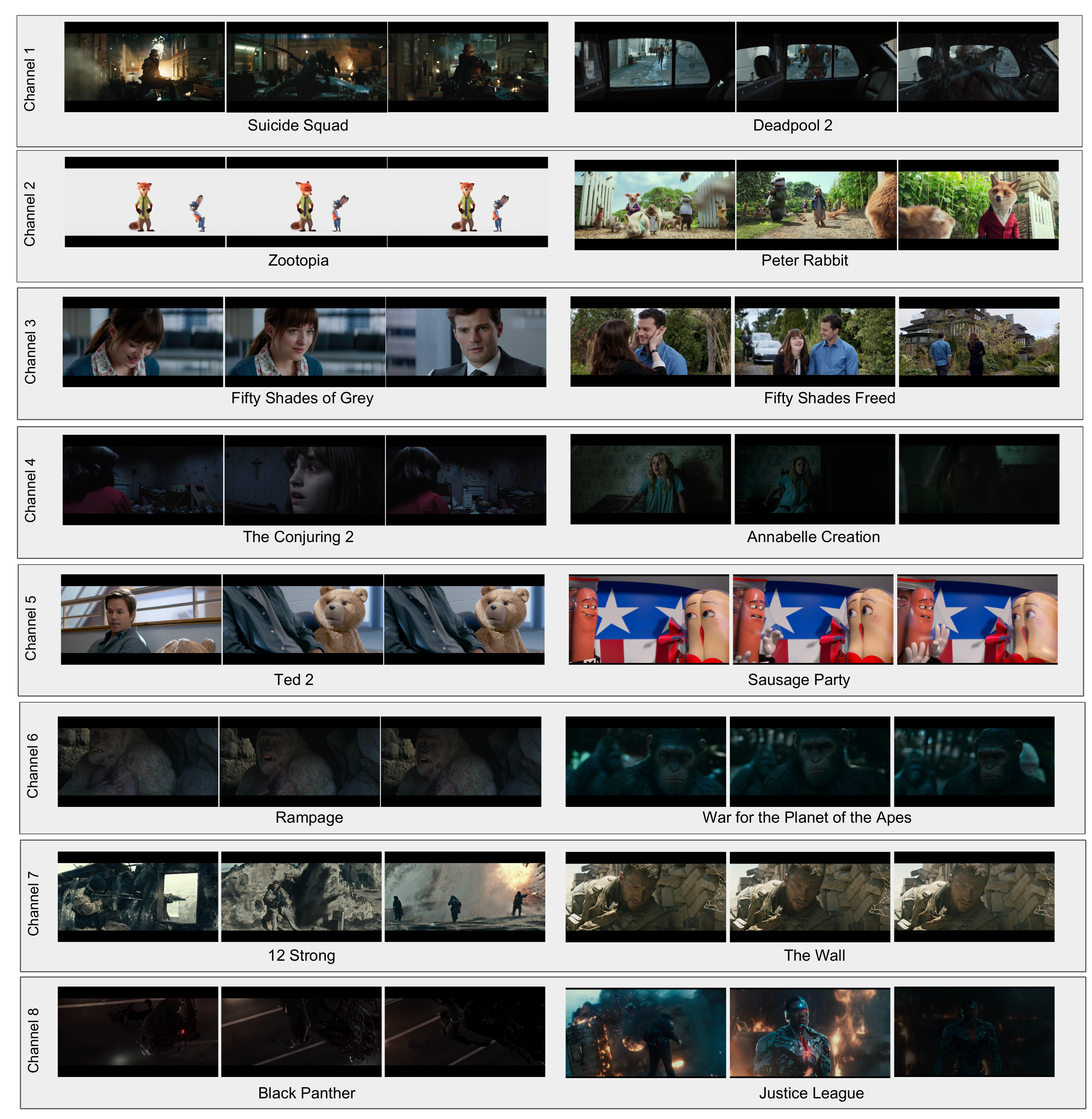}
  \caption{Each row shows two sample video sequences from different trailers that highly \textit{activate} a particular channel at the last \textbf{relu} layer (before entering the average pooling layer). These video sequences result in the activation that is at least two standard deviation higher than the average activation at their corresponding channels. These example suggests that the proposed video convolutional model captures scenes and actions across a wide variety of genres, including action, animation, romance, horror, monster, and war.}
  \label{fig:samples}
\end{figure*}

\subsection{Model Explainability}
The activation of the convolutional filters in the proposed video convolutional model provides insights into the \textit{object-sequence templates} that the model actually learns. This is not possible with an AvgPool method where frame-features are collapsed. 

Figure \ref{fig:samples} shows sample actual video sequences that \textit{highly activate} different channels at the last \textbf{relu} layer (before entering the average pooling layer). The examples in figure 3 suggest that the proposed video convolutional model does a good job capturing common object-sequences across wide variety of genres, including action, animation, romance, horror, monster, and war. In the examples, for instance, Channel 4 appears to be activated by the slow and intense close-up shots that are typical in horror movies, whereas Channel 8 appears to perhaps be activated by stunt performances that include flipping vehicles that are typical in action movies or superhero movies.

%


\section{Discussion}
This work presents the analysis of temporal sequences of objects that appear in movie trailers when applied to the problem of movie recommendation and audience prediction. 

Movie trailers are engineered to increase awareness and to present key aspects of the story and the cinematography. Filmmakers and studios use movie trailers to increase the urge to see the movie. To do that, they rely on proven techniques and templates to create trailers that maximize intent to watch while at the same time ensuring that the unique elements of the story--what makes the movie worth seeing--get appropriately reflected in the trailer. Every trailer is different, but there are commonalities between trailers of movies that belong to the same genre. 

The temporal sequencing of elements in a movie trailer (e.g., when to introduce a character, for how long, etc) is an aspect that filmmakers and studios pay careful attention to because of the short format of the trailer. As with other aspects of the trailer, temporal sequencing follows norms that are tried and tested. Moreover, the templates used for different types of movies are also different. Moviegoers decisions about which movies to go see can be projected on the feature space of movie trailers to create an implicit measure of similarity or dissimilarity between trailers. A moviegoer that buys tickets to a movie has probably seen movies in the past the trailers of which contained similar sequences. When we characterize trailers using object-sequences, as we do here, moviegoers actions are implicitly telling us which object-sequences matter to measure trailer similarity.

Our results show that recommendation systems that are based on the analysis of object-sequences have more predictive power in cold start situations than systems based on average pooling of video frames. Object-sequences are more effective at predicting customer behavior because they provide a more efficient way to represent trailers than simple average pooling. They are more efficient because they use convolutional filters to learn which distinct temporal sequences are optimal for each of the 1024 dimensions of the frame embeddings, and use collaborative filters to learn which non-linear combination of sequences is optimal for prediction. The resulting convolutional and collaborative network architecture contributes not only to isolate the specific components of the video signal that are more helpful for the prediction problem, but to increase the explain-ability of the model predictions.

\bibliographystyle{plain}
\bibliography{references}

\end{document}